\documentclass[letterpaper, 10 pt, journal, twoside]{IEEEtran}
\usepackage{cite}
\usepackage[hidelinks]{hyperref}
\usepackage{xcolor}
\usepackage[pdftex]{graphicx}
\usepackage{amsmath, amsfonts}
\usepackage{algorithm}
\usepackage{algorithmic}
\usepackage{array}
\ifCLASSOPTIONcompsoc
 \usepackage[caption=false,font=normalsize,labelfont=sf,textfont=sf]{subfig}
\else
 \usepackage[caption=false,font=footnotesize]{subfig}
\fi
\usepackage{stfloats}
\ifCLASSOPTIONcaptionsoff
 \usepackage[nomarkers]{endfloat}
\let\MYoriglatexcaption\caption
\renewcommand{\caption}[2][\relax]{\MYoriglatexcaption[#2]{#2}}
\fi
\usepackage{url}
\begin{document}

\title{Learn to Teach: Sample-Efficient Privileged Learning for Humanoid Locomotion over Real-World Uneven Terrain}

\author{Feiyang Wu$^{1}$, Xavier Nal$^{2}$, Jaehwi Jang$^{1}$, Wei Zhu$^{1}$, Zhaoyuan Gu$^{1}$, Anqi Wu$^{1}$, Ye Zhao$^{1}$%
\thanks{Manuscript received: March, 12, 2025; Revised Jun, 11, 2025; Accepted July, 12, 2025.}%
\thanks{This paper was recommended for publication by Editor Abderrahmane Kheddar upon evaluation of the Associate Editor and Reviewers' comments.
This work was supported in part by the Office of Naval Research (ONR) under Grant N000142312223; in part by the National Science Foundation (NSF) under Grant IIS-1924978, Grant CMMI-2144309, and Grant FRR-2328254; and in part by United States Department of Agriculture (USDA) under Grant 2023-67021-41397.} 
\thanks{$^{1}$Feiyang Wu, Jaehwi Jang, Wei Zhu, Zhaoyuan Gu, Anqi Wu, Ye Zhao are with Georgia Institute of Technology, Atlanta, USA.
        {\tt\footnotesize {\{feiyangwu, jjang318, wzhu328, zgu78, anqiwu, yezhao\}@gatech.edu}}}%
\thanks{$^{2} $Xavier Nal is with the School of Engineering, \"{E}cole Polytechnique Fédérale de Lausanne, 1015 Lausanne, Switzerland. {\tt\footnotesize xavier.nal@alumni.epfl.ch}}%
\thanks{Digital Object Identifier (DOI): see top of this page.}
}

\markboth{IEEE Robotics and Automation Letters. Preprint Version. Accepted July, 2025}
{Wu \MakeLowercase{\textit{et al.}}: Privileged Learning for Humanoid Locomotion}

%



\maketitle

\begin{abstract}
Humanoid robots promise transformative capabilities for industrial and service applications. While recent advances in Reinforcement Learning (RL) yield impressive results in locomotion, manipulation, and navigation, the proposed methods typically require enormous simulation samples to account for real-world variability. This work proposes a novel one-stage training framework—Learn to Teach (L2T)—which unifies teacher and student policy learning. Our approach recycles simulator samples and synchronizes the learning trajectories through shared dynamics, significantly reducing sample complexities and training time while achieving state-of-the-art performance. Furthermore, we validate the RL variant (L2T-RL) through extensive simulations and hardware tests on the Digit robot, demonstrating zero-shot sim-to-real transfer and robust performance over 12+ diverse terrains without depth estimation modules. Experimental videos are available at \href{https://lidar-learn-to-teach.github.io}{https://lidar-learn-to-teach.github.io}.
\end{abstract}


%
\IEEEpeerreviewmaketitle

\section{Introduction}
\IEEEPARstart{R}{einforcement} Learning (RL) has revolutionized robotic control by tackling complex tasks such as dynamic locomotion \cite{Siekmann2021Stair, lirobust2023, haarnoja2023learning}. Despite these achievements, policies trained in simulators often falter when deployed into the real world due to the inevitable simulation-to-reality gap \cite{kober2013reinforcement}. Although domain randomization \cite{peng2018sim} is widely used to mitigate these discrepancies, it incurs significantly higher sample complexity as agents must explore extensive environmental variations.

Recently, teacher-student learning methods have demonstrated promising results by leveraging an expert teacher to guide students with restricted observation spaces \cite{chen2020learning, lee2020learning, miki2022learning}. However, the conventional two-stage training discards valuable teacher interactions with the environment and often suffers from mismatches between independently trained teachers and students.
To address these issues, we propose Learn-to-Teach (L2T): a unified training framework that co-trains teacher and student agents in a single, interactive stage, where the student fully utilizes the collected samples.  

\begin{figure*}
    \vspace{0.2cm}
    \centering
    \includegraphics[width=0.9\textwidth]{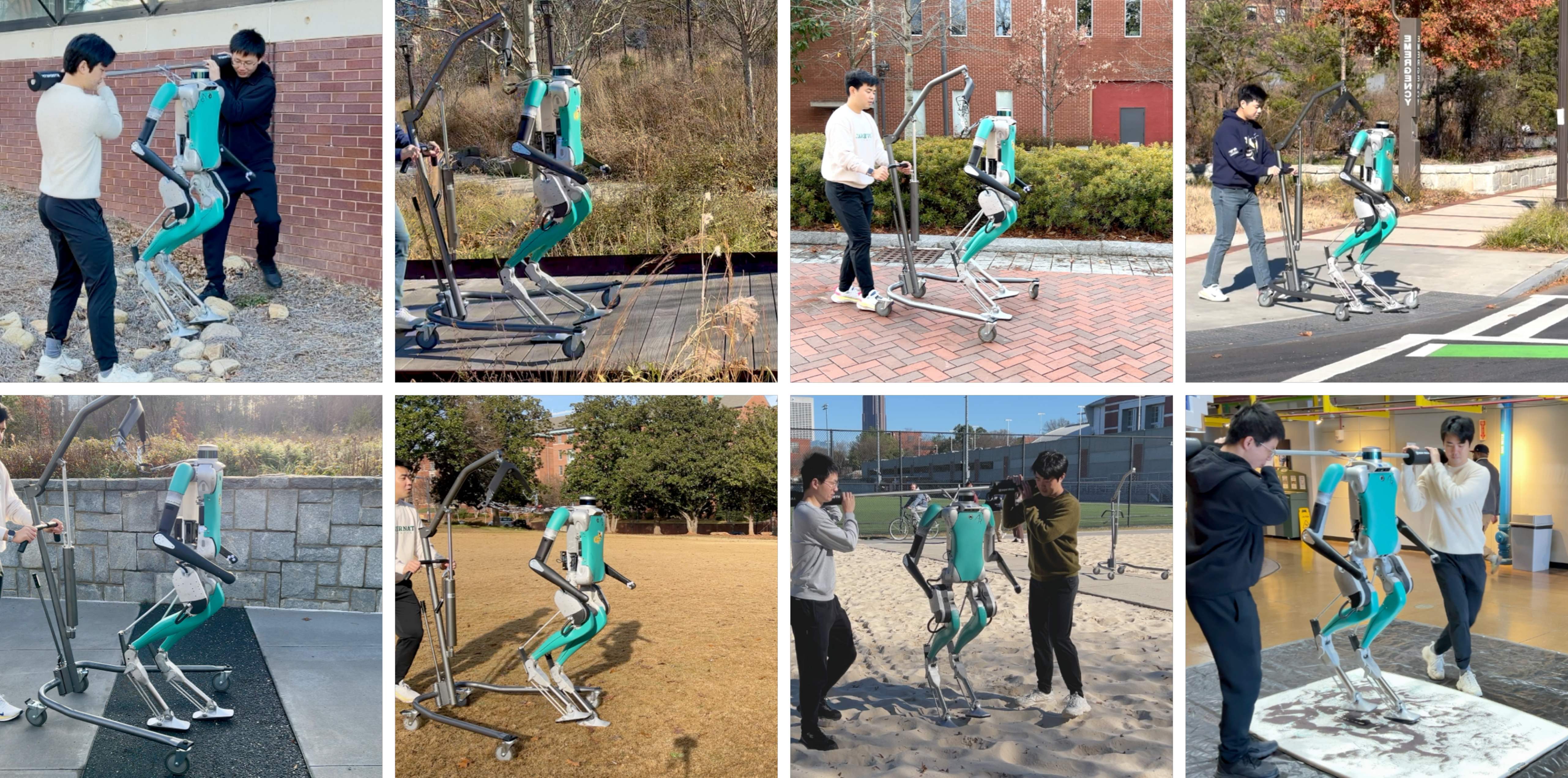}
    \caption{We implement our L2T-RL algorithm on our bipedal walking robot Digit and deploy it on diverse terrain with various environmental conditions such as wet grass, gravel, sandy terrain, and slippery surfaces. }
    \label{fig:digit_on_terrains}
    \vspace{-0.3cm}
\end{figure*}

To quantify L2T’s advantages, we implement L2T-RL, an RL variant, and benchmark its performance on humanoid locomotion tasks using the Digit robot in Isaac Lab—a state-of-the-art GPU-accelerated simulator \cite{mittal2023orbit}.
Our results show that L2T-RL can achieve stable and superior performance compared to the conventional teacher-student learning paradigm, requiring $50\%$ fewer samples. 
Consequently, we deploy our trained policy on the robot Digit and conduct extensive hardware experiments in indoor and outdoor environments. 
Strikingly, the resulting student agent, a lightweight LSTM-based policy, exhibits zero-shot sim2real transfer on the physical Digit robot across a wide range of terrains, including gravel, sand, grass, and slopes (Fig.~\ref{fig:digit_on_terrains}).
We also test our control policy on various perturbations such as push recovery, walking under payload, and walking on slippery or wet terrains or with the wind blowing (see Fig.~\ref{fig:perturbation_tests} and the supplementary video). 
Our contributions are as follows:

\textbf{Efficient training framework}: 
We propose a joint teacher-student training paradigm that optimizes both policies simultaneously. 
Unlike prior decoupled approaches, our framework enables cross-agent knowledge transfer to the student policy by 
dynamically utilizing the teacher’s training samples directly within a single training stage, avoiding the need for training from scratch in a separate stage.

\textbf{Mitigation of teacher-student imitation gap}: We propose a sample mixing strategy to alleviate the imitation gap between the teacher and student, which traditional privileged learning is unable to address \cite{weihs2021bridging}. Both agents will contribute to the replay buffer following a predefined schedule when collecting samples. Mixing samples enables a joint optimization process that mitigates the imitation gap while promoting sample efficiency by letting both agents explore Out-of-Distribution (OOD) data.

\textbf{Humanoid RL agent deployment}: 
We demonstrate real-world locomotion agility through hardware experiments. Our policy, trained entirely in simulation, enables a physical humanoid robot to reliably traverse $12+$ real-world terrains (concrete, gravel, slopes, stairs, etc.) and withstand dynamic perturbations (pushes and payloads) without offline fine-tuning. The policy achieves high success in unstructured environments, matching the teacher’s robustness despite using only proprioceptive inputs without depth estimation modules.

\section{Related work}
\textbf{Teacher-student learning}:
    In the robotics learning community, teacher-student learning \cite{chen2020learning, lee2020learning, miki2022learning} has gained significant attention due to its applicability and effectiveness in addressing sim2real challenges. 
    In this framework, the teacher agent is trained with complete knowledge of the state space. 
    After obtaining an expert-level teacher, a student agent is trained in an observation space that follows the available sensor configurations on hardware, where the goal is to imitate the teacher's action \cite{miki2022learning}. 
    In this work, we extend this learning framework by training the teacher and the student simultaneously in a single stage. 
    Prior work \cite{wang2024cts} proposed a method termed Concurrent Teacher Student (CTS) learning, which also explored the idea of co-training both agents. However, CTS trains a shared policy across the teacher and the student, only differentiating the observation encoder and the critic. 
    This potentially disrupts the training process as the privileged critic naturally rewards actions that might not seem valuable to the student.
    In comparison, L2T trains two separate agents, with the option of sharing an encoder network or using an asymmetric learning style critic. 

\textbf{Learning with partial observation}: 
    Recent advancements in RL under partial observability have significantly improved the ability of robotic systems to operate in complex, uncertain environments \cite{miki2022learning}. Contemporary approaches often leverage deep recurrent architectures, such as \cite{ni2021recurrent}, to infer latent state representations from sequential data, effectively bridging traditional POMDP solvers with modern deep RL frameworks. In robotics control, practitioners construct history-dependent policies from a sliding-window style observation or rely on the recurrent architecture of the policy network. At the same time, asymmetric learning has emerged as another effective strategy to bridge the gap between training and execution \cite{pinto2017asymmetric}. In these approaches, the critic network is provided access to privileged, full-state information during training. Recent works have demonstrated that such asymmetric actor-critic frameworks improve sample efficiency and enhance policy robustness \cite{ma2023learning}. In this work, we combine these learning techniques, utilizing a recurrent network and an asymmetric critic, to solve the underlying POMDP problem efficiently.

\textbf{Learning from demonstrations}:
    Learning from demonstrations (LfD) has attracted significant interest in the robot learning field due to the growing abundance of robot data and the popularity of simple yet effective imitation learning (IL) frameworks \cite{o2024open}. 
    LfD has demonstrated impressive results in controlling robot manipulators for tabletop tasks \cite{ze2024generalizable}.
    A recent surge of LfD studies in humanoids and bipeds have shown the promising potential of whole-body control and loco-manipulation \cite{cheng2024expressive, gu2025humanoid, ben2025homie, ze2024generalizable}. 
    However, supervised learning demands high-quality behavior data, oftentimes through elaborate data collection pipelines \cite{ben2025homie, ze2024generalizable} and/or needs accurate re-targeting to robot states from datasets with different morphologies
    On the other hand, the prevalent IL loss is known to be suboptimal from a learning perspective \cite{wu2023inverse}. Thus, in this work, we focus on a generic algorithm framework to address the sim2real gap alone, without the interference of possible issues brought up by LfD methods. Furthermore, our proposed framework can be easily extended to the LfD setting, which we leave as a future direction.

\begin{figure*}
\vspace{0.2cm}
  \centering
  \includegraphics[width=0.9\linewidth]{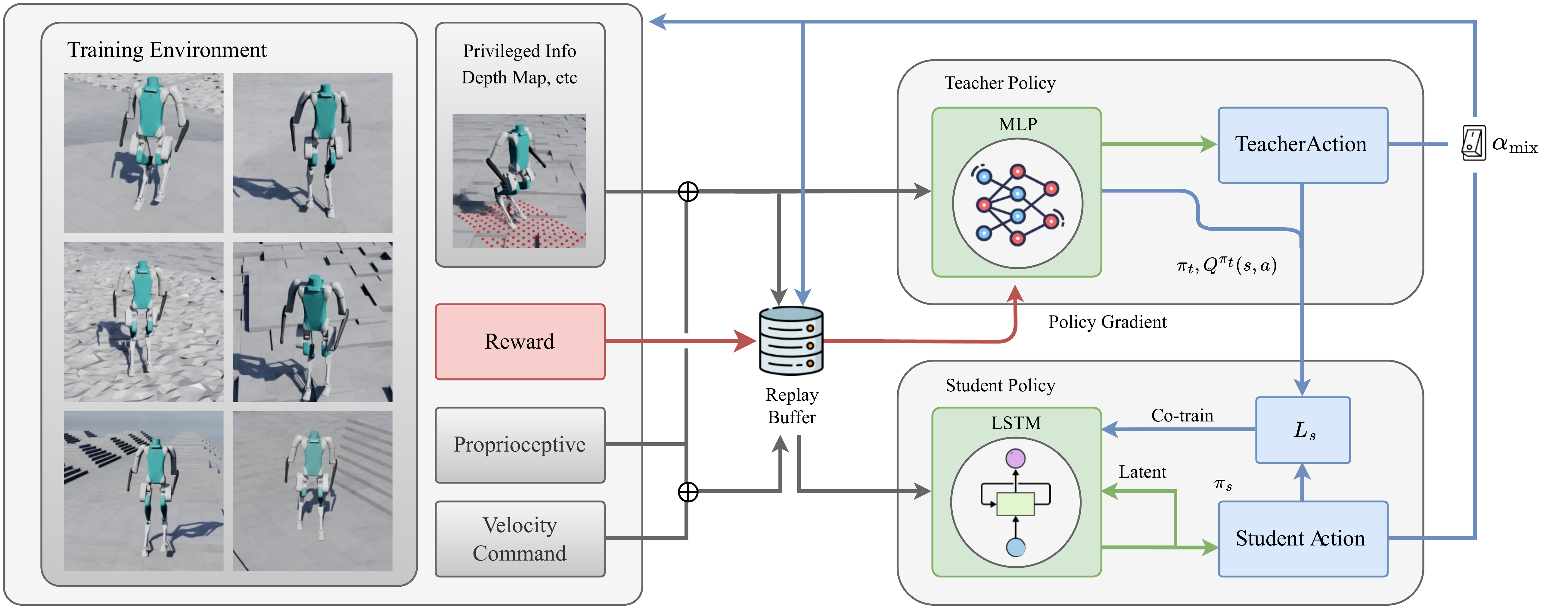}
  \caption{Learn to Teach (L2T) training pipeline. The teacher agent utilizes a neural network for the policy, which comprises three fully connected layers with sizes $[512, 256, 128]$. The student agent's policy network deployed on the robot is an LSTM network with a hidden layer of $128$ units, followed by fully connected layers with shape $[512, 256, 128]$. The teacher learns via conventional RL methods, while the student updates its policy by imitating the teacher. }
  \label{fig:l2t_training_pipeline}
  \vspace{-0.3cm}
\end{figure*}

\textbf{Bipedal locomotion over complex terrain}: 
    Humanoid robots recently have gained increasing interest due to their applicability and versatility \cite{gu2025humanoid,chen2024learning, dugar2024learning, cheng2024expressive, gu2024humanoid, long2024learning, ben2025homie, wu2024infer}, 
    ranging from locomotion \cite{radosavovic2024real}, to manipulation \cite{liu2024opt2skill}.
    Prior bipedal locomotion works \cite{radosavovic2024real, radosavovic2024humanoid} have explored the conventional teacher-student learning paradigm in locomotion tasks. 
    However, the training process can take significant samples even with an elaborate training environment design. 
    Concurrent works also incorporate memory structure into the policy architecture \cite{wang2024cts, radosavovic2024real, dao2024sim}, or learning from demonstrations collected from various data sources such as human motion \cite{radosavovic2024learning} or generation using model-based methods \cite{liu2024opt2skill}. 
    In comparison, we design straightforward and intuitive reward functions for bipeds in general terrain settings, offering a simple yet effective solution.

\section{Methods} \label{sec:methods}
This section introduces our problem setup and notations and presents our learning framework. 
A Markov Decision Process (MDP), denoted as $\mathcal{M}$, is described by a tuple: $\mathcal{M}=\langle\mathcal{S}, \mathcal{A}, R, \mathcal{P}, \Pi, \gamma \rangle$, where an agent starts with a given state $s_0$ following the initial state distribution $p(s_0)$. 
At any time step $t$, the agent at the current state $s_t\in \mathcal{S}$ takes an action $a_t \in \mathcal{A}$ following the agent’s policy $\pi\in \Pi$, which defines a probability distribution over action space for each state. 
While receiving an instantaneous scalar reward $r(s_t, a_t) \in \mathbb{R}$, the state of the agent transitions to a new state $s_{t+1}\in \mathcal{S}$ following a transition model $\mathcal{P}(\cdot|s_t,a_t)$. 
$\gamma$ specifies the discount factor. 
The goal of the agent is to maximize the expected discounted sum of rewards the agent receives over time
$\max_{\pi} \mathbb{E} \left[ \sum^{\infty}_{t=0} \gamma^t r(s_t, a_t)\right]
$, where the expectation is taken over actions $a_t\sim \pi(\cdot| s_t)$, and transition probabilities $s_{t+1} \sim \mathcal{P}(\cdot| s_t, a_t)$ and initial distribution $s_0\sim p(s_0)$.
A Partially Observable Markov Decision Process (POMDP) is further coupled with an observation model $O(\cdot|s_t)$, which is generally hidden from the agent. At each time step, the agent only observes $o_t\sim O(\cdot|s_t)$ sampled from the observation model. Then, the agent takes an action based on $o_t$, following its policy $\pi(\cdot|o_t)$, and subsequently receives a reward from the environment $r(s_t, a_t)$.
To train a policy robust to various observation models, 
domain randomization is often applied. For example, by adding noise to the state $s_t$, the trained agent’s policy can handle observations $o_t=s_t+\epsilon$, where $\epsilon$ can be any noise distribution. 

We train the teacher with a generic actor-critic method. During training, the teacher interacts with the environment, generates samples, and stores them in a replay buffer \cite{lin1992self}. 
In standard teacher-student frameworks, the teacher’s collected samples are used solely for training the teacher policy $\pi_t$ and then discarded. In contrast, our L2T framework co-trains the student with the teacher, reusing the teacher’s samples across all iterations. Fig.~\ref{fig:l2t_training_pipeline} illustrates our learning framework, and Algorithm~\ref{alg:learn_to_teach_rl} presents the pseudo-code. We employ an MLP for the teacher because, given privileged access to the full state (depth scans, root pose, terrain profile, etc.), each observation is already fully descriptive and Markovian so that there is no need to model temporal dependencies via recurrence. We also notice that this architecture has been widely adopted by concurrent teacher-student learning paradigms \cite{radosavovic2024real}. Specifically, as the teacher interacts with the environment, we record samples \((s, a, r, s')\), where $s'$ denotes the next observation, the corresponding noisy observations \(o\) generated by domain randomization, and $o'$, the next observation. In other words, we store \( (s, o, a, r, s', o')\) as training data in the replay buffer. 
The student updates its policy at each iteration by sampling mini-batches from the replay buffer, but its policy relies solely on the collected noisy data. This joint training procedure greatly improves sample efficiency as both agents learn together, without the need for a separate stage as used in the traditional set-up.

Another key challenge in the teacher-student framework is the discrepancy between the teacher’s and the student’s observation spaces. 
Traditional teacher-student learning methods fall short because the teacher does not account for the limitations of the student’s observations, leading to suboptimal guidance \cite{weihs2021bridging}. 
To bridge this gap, we introduce a sample-mixing mechanism in which the student collects its own samples directly from the environment. 
These student-generated samples—including actions and the resulting observations—are incorporated into the replay buffer as if they were produced by the teacher. 
This injection of OOD data helps reduce the imitation gap between the two agents.

To systematically blend teacher and student experiences, we define a mixture coefficient \(\alpha_{\text{mix}}\). At each time step, the action \(a\) is determined by a probabilistic mixture of the teacher’s policy \(\pi_t\) and the student’s policy \(\pi_s\). 
As such, we collect trajectories \(s_1, a_1, s_2, a_2, s_3, a_3\) where \(a_i\) could either come from the teacher or the student. This sample mixing mechanism ensures that information from the student agent back propagates to the teacher.
Specifically, the action selection is defined as follows:
\begin{equation} \label{eq:scheduling}
a =
\begin{cases}
\text{sample } \pi_s(\cdot \mid o), & \text{with probability } \alpha_{\text{mix}}, \\
\text{sample } \pi_t(\cdot \mid s), & \text{with probability } 1 - \alpha_{\text{mix}}.
\end{cases}
\end{equation}
Additionally, \(\alpha_{\text{mix}}\) is scheduled linearly from 0 to a predefined constant over the course of training, which is $0.2$ in our implementation. This formulation ensures that, initially, the teacher’s guidance dominates the action selection, but as training progresses, the student’s policy increasingly influences the learning. The scheduling of \(\alpha_{\text{mix}}\) helps balance the contributions of both policies and guarantees stable training.
By combining these strategies, our framework leverages both the teacher’s guidance and the student’s explorative capabilities to achieve more robust learning outcomes.
This is in steep contrast to CTS, where the teacher encoder and the student encoder trains on independent datasets.

We implement a variant of our framework, L2T-RL. 
We apply policy gradient methods to update the teacher policy $\pi_t$. At each iteration $k$, the critic is updated by estimating the value functions: \(V^{\pi}(s):= \mathbb{E}_{\pi}\left[\sum_{t=0}^\infty \gamma^t\,r(s_t,a_t) \,|\, s_0=s\right],\\
  Q^{\pi}(s,a) := \mathbb{E}_{\pi}\left[\sum_{t=0}^\infty \gamma^t\,r(s_t,a_t) \,|\, s_0=s, a_0=a\right], \)
where $V^{\pi}(s)$ is the value function, and $Q^{\pi}(s,a)$ is the discounted action-value function. For brevity, we will only use the subscript $t$ to denote the teacher from now on and use subscript $s$ to denote the student.
Subsequently, the teacher’s policy is updated via a Policy Mirror Descent \cite{lan2023policy} step with a step size $\beta$:
\begin{equation} \label{eq:teacher_policy_update}
  \min_{p_t}\; -\beta \langle Q^{\pi_t}(s,\cdot), p_t(\cdot|s) \rangle + \text{KL}(\pi_t\|p_t)\quad \forall s\in \mathcal{S},
\end{equation}
where the optimal $p_t$ represents the teacher policy in the next iteration.
Any policy improvement scheme can be fit into the L2T framework. 
This formulation encompasses a range of policy gradient methods \cite{lan2023policy}, such as Proximal Policy Optimization (PPO) \cite{schulman2017proximal} and Soft Actor-Critic (SAC) \cite{haarnoja2018soft}. In our experiments on the Digit robot, we employ a PPO-style update. However, our framework can be easily extended to various learning methods, including imitation learning methods, or Inverse Reinforcement Learning (IRL) methods such as in \cite{wu2023inverse}.
In practice, we pool samples from both agents to update the teacher policy using PPO. If we can maintain \(\pi_s\approx\pi_t\), any student action \(a_s \sim \pi_s(\cdot|s)\) satisfies \(\pi_t(a_s| s)>0\), allowing us to view these student-generated transitions as valid (albeit lower-probability) samples from the teacher’s distribution.

For the student policy, we consider two choices for loss functions. First, we can minimize an imitation loss between the teacher’s and the student’s policies:
\begin{equation} \label{eq:bc_loss}
  \min_{p_s}\; L_{\text{IL}} = \mathbb{E}_{s,\, o \sim D} \left\| p_s(\cdot|o) - \pi_t(\cdot|s)\right\|_2,
\end{equation}
where $D$ denotes the replay buffer and the optimal $p_s$ represents the student policy in the next iteration. Alternatively, one may minimize the KL divergence between the two:
\begin{equation} \label{eq:kl_loss}
  \min_{p_s}\; L_{KL} = \mathbb{E}_{s,\,  o\sim D}\; \text{KL}\left(p_s(\cdot|o)\;\|\;\pi_t(\cdot|s)\right),
\end{equation}
or any statistical distance metric that fits the action space. 

Besides the imitation loss, the student can be updated using an asymmetric learning approach \cite{pinto2017asymmetric} that leverages the teacher’s critic, i.e., the value functions:
\begin{equation} \label{eq:asym_loss}
  \min_{p_s} L_{\text{Asym}} = -\beta \langle Q^{\pi_t}(s,\cdot), p_s(\cdot|o) \rangle + \text{KL}(p_s\| \pi_s).
\end{equation}
We denote the general loss function for student agents as $L_s$. In our application on the Digit robot, we observed that using the $L_\text{IL}$ imitation loss yields the best performance, while the addition of $L_{\text{Asym}}$ does not affect the overall performance by a large margin. We conjecture that the $L_\text{IL}$ loss allows the student policy to have a slightly higher exploration capability as we observe that $L_\text{Asym} + L_\text{IL}$ will reach a training plateau that is inferior in performance than using $L_\text{IL}$ alone.

\begin{algorithm}
  \caption{Learn to Teach - RL (L2T-RL)}
  \label{alg:learn_to_teach_rl}
  \begin{algorithmic}[1]
    \REQUIRE initial teacher policy $\pi_t^0$, student policy $\pi_s^0$, and step size sequences $\{\beta^k_t\}$ and $\{\beta^k_s\}$
    \FOR{$k = 0$ \TO $K$}
    \STATE Sample a mini-batch $D_k$ from the replay buffer $D$
    \STATE Update the teacher critic
    $Q^{\pi_{t}^{k+1}}(s,a), V^{\pi_t^{k+1}}(s)$
    \STATE Update the teacher policy:
    \[
    \pi^{k+1}_t = \arg\min_{p_t} \left[ -\beta^k_t \langle Q^{\pi_t^k}(s,\cdot), p_t(\cdot|s) \rangle + \mathrm{KL}(\pi_t^k \| p_t) \right] 
    \]
    \vspace{-0.3cm}
    \STATE Update the student policy:
    \[\pi^{k+1}_s = \arg\min_{\pi_s} L_{s}(\pi_s^k)\]
    \vspace{-0.3cm}
    \STATE Roll out to collect new samples $D'$ according to the scheduling in Eq.~\ref{eq:scheduling}
    \STATE Update the replay buffer: $D \leftarrow D \cup D'$
  \ENDFOR
  \end{algorithmic}
\end{algorithm}
\vspace{-0.5cm}

\section{Environment Design}
The Digit robot is a bipedal walking robot with $30$ degrees of freedom, which includes $20$ actuated joints with $4$ per arm and $6$ per leg. All joints are revolute joints except for the shin and heel joints, which are spring-based. Notably, the Digit robot features three closed kinematic chains per leg. Two of these chains involve motors controlling the foot, assisted by additional rods, while the third chain is responsible for controlling the heel via a rod extending from the hip. This leg design makes it a significantly challenging task for RL algorithms due to the high-dimensional action space and the complex dynamics of the robot. 
We highlight a significant portion of our work is to reconstruct a faithful Universal Scene Description (USD) model of the robot in IsaacLab  \cite{mittal2023orbit}, although this is not claimed as an algorithmic contribution. As a result, we build a velocity-tracking RL task with accurate dynamics w.r.t the robot hardware.

\subsection{Observation space} The observation space (see table \ref{table:observation_terms}) is constructed using data provided by the robot’s sensors, including base linear velocity, base angular velocity, joint positions, and joint velocities. Additionally, we include the commanded velocity that the robot will receive from an external controller during execution, the computed projected gravity based on the IMU data, and desired gait phase based on the robot execution time. Finally, the actions in a previous time step are also incorporated into the observation space, which allows us to learn a history-dependent policy using recurrent neural nets. We model measurement noise as $o_t = s_t + \alpha \epsilon$ where $\alpha$ is the scale, and $\epsilon$ is either Gaussian or uniform noise. We add this noise to the student's observation space to mimic the hardware sensors while keeping the teacher's observation noise-free, except for the ones that incorporate the student's observations in order to alleviate the imitation gap, which is considered a common practice. Additionally, privileged information (lower half of Table~\ref{table:observation_terms}) is provided for the teacher for easier training. 
While our locomotion experiments randomize only the base mass, we include all environment parameters as privileged information, bundling them simplifies our environment API for future extensions. Note that the base linear velocity is given by the low-level software APIs provided by Agility Robotics.
\vspace{-0.5cm}
\begin{table}[htbp]
\centering
\caption{Observation Terms for Teacher and Student}
\label{table:observation_terms}
\begin{tabular}{ccccc}
\hline
\textbf{Observation Terms} & \textbf{Dim} & \textbf{Noise} & \textbf{Student $\pi_s$} & \textbf{Teacher $\pi_t$} \\
\hline
Clock input                  & 2   &               & \checkmark & \checkmark \\
Base lin. vel.           & 3   & \checkmark  & \checkmark & \checkmark \\
Base ang. vel.           & 3   & \checkmark  & \checkmark & \checkmark \\
Projected gravity        & 3   & \checkmark  & \checkmark & \checkmark \\
Velocity command         & 3   &               & \checkmark & \checkmark \\
Joint pos.               & 30  & \checkmark  & \checkmark & \checkmark \\
Joint vel.               & 30  & \checkmark  & \checkmark & \checkmark \\
Last action              & 20  &               & \checkmark & \checkmark \\ 
\hline
Root state (w)           & 7   &               &            & \checkmark \\
Base lin. vel. (w)       & 3   &               &            & \checkmark \\
Base ang. vel. (w)       & 3   &               &            & \checkmark \\
Base pos. (w)            & 3   &               &            & \checkmark \\
Base quant. (w)          & 4   &               &            & \checkmark \\
Env params               & 316 &               &            & \checkmark \\
Height scan              & 187 &               &            & \checkmark \\
\hline
\end{tabular}
\vspace{-0.3cm}
\end{table}

\subsection{Action space} The action space is designed as the target full-body joint positions $q_{\rm target}$, which a Proportional Derivative (PD) controller will aim to track during execution. At a frequency of $50$ Hz, the policy predicts the current targeted joint based on the current observation, and then at a higher frequency ($1$ kHz), the PD controller computes the torque $\tau$ as inputs to the motors to control the robot’s joints. The target velocity is set to zero, which is commonly employed in legged robot research. The PD gains are determined through empirical tuning to ensure stable joint control. We use a standard PD control law for computing the torque, i.e, \(\tau = \text{K}_p (q_{\text{target}} - q) + \text{K}_d (\dot{q}_{\text{target}} - \dot{q}) \), where $q$ represents the measured joint positions and $\dot{q}$ represents the measured joint velocities.

\subsection{Reward functions} Our reward function design is summarized in Table~\ref{table:reward_functions}. We adopt some of the existing reward functions in IsaacLab across other velocity command tasks for bipeds and quadrupeds, including termination penalty, action rate, joint deviation, etc. In addition, we design specific reward functions for training the Digit robot, for which we highlight two of them. For implementation details, please refer to our code.
\begin{table}[t]
\vspace{0.1cm}
\centering
\caption{Reward Functions and Their Weights}
\label{table:reward_functions}
\begin{tabular}{cc|cc}
\hline
\textbf{Reward}       & \textbf{Weight} & \textbf{Reward} & \textbf{Weight} \\ \hline
Termination penalty            & -200.0        & Foot contact             & 2.0           \\
Being alive                    & 0.01          & Track foot height        & 0.5           \\ 
Action rate                    & -0.015        & Foot clearance           & 0.5           \\ 
DOF velocity                   & -5e-4         & Track lin vel XY         & 0.5           \\ 
Undesired contacts             & -1.0          & Track ang vel Z          & 1.0           \\ 
Flat orientation               & -10.0         & Lin vel XY               & -2.0          \\ 
Feet air time                  & 0.25          & Ang vel Z                & -0.1          \\ 
Feet sliding                   & -1.0          & DOF torques              & -1.0e-5       \\ 
DOF pos limits                 & -0.5          & DOF acc                  & -2.5e-7       \\ 
Joint deviation hip            & -5.0          & Joint deviation toes            & -0.1         \\ 
Joint deviation arms           & -0.3          &              &                 \\  \hline
\end{tabular}
\vspace{-0.5cm}
\end{table}

\textbf{Track Foot Height:} 
  We reward the agent for following a desired foot height trajectory, which is precomputed as a quintic polynomial.
  \begin{align}
    r_{\mathrm{foot-track}} &= \exp\{ -\| h_\text{foot\_traj} - h_\text{foot\_traj\_target}\|_2\},
  \end{align}
  $h_\text{foot\_traj}$ is the actual foot height and $h_\text{foot\_traj\_target}$ is the target foot height. We adjust the desired foot height based on the current CoM position to adapt to uneven terrains. Specifically, instead of tracking the absolute height of the foot, we track the relative distance of the foot and the CoM position to compensate for the terrain height, which is hard to obtain at run time. This reward produces significantly different motions than rewarding foot clearance alone and it is one of the most important reward term, without which we are unable to train an agent robust to uneven terrains.
  Note that, this reward is different from the foot clearance reward as the latter rewards the foot to reach a desired height relative to floor. On flat ground, the desired height is consistent with the peak of the reference foot trajectory.

\textbf{Foot Contact Matching:} 
  \begin{equation}
    r_{\mathrm{contact}} =
    \begin{cases}
        c_1, & \text{sign}(\phi(t)) = \text{sign}(F_{\text {GRF}}>0) \\ -c_2 ,&  \text{otherwise.}
    \end{cases}
  \end{equation}
  where $\phi(t) = \sin(2\pi t/h)$, $h=0.68$ is the gait cycle duration, $c_1, c_2\in\mathbb{R}^+$ are constants. 
  We reward the agent when the foot contact matches the desired gait cycle. 
  For example, if $\phi(t) > 0$ (indicating the foot should be in contact with the ground) and $F_{\text {GRF}}>0$ (indicating the foot is actually in contact), we assign a positive reward $c_1$. Otherwise, a penalty $-c_2$ is applied.

\vspace{-0.3cm}
\begin{table}[ht]
\caption{Event Terms for Domain Randomization}
\centering
\begin{tabular}{l|l}
\hline
\textbf{Name}  & \textbf{Name}  \\ \hline
Rand. friction coeff & Add base mass  \\
Rand. gravity  & Add external force  \\
Rand. base location  & Push robot  \\ 
Rand. robot joints  &  \\
\hline
\end{tabular}
\label{table:event_terms}
\vspace{-0.5cm}
\end{table}

\subsection{Domain randomization and curriculum} We dynamically change the observation space and perturb the physical dynamics of the environment in the hope of capturing the randomness and variation of the real world. 
We list all the domain randomization schemes used in the training in Table.~\ref{table:event_terms}. 
In practice, we find that randomizing friction coefficients and adding external pushes can greatly improve the robustness of the trained policy.

Additionally, we adopt the curriculum training setup implemented by IsaacLab.
We utilize the existing terrain map, which includes seven different terrains: flat ground, slopes, stepping stones, and pyramid stairs up and down. 

\begin{figure*}[ht]
    \centering
    \vspace{0.2cm}
    \includegraphics[width=0.95\linewidth]{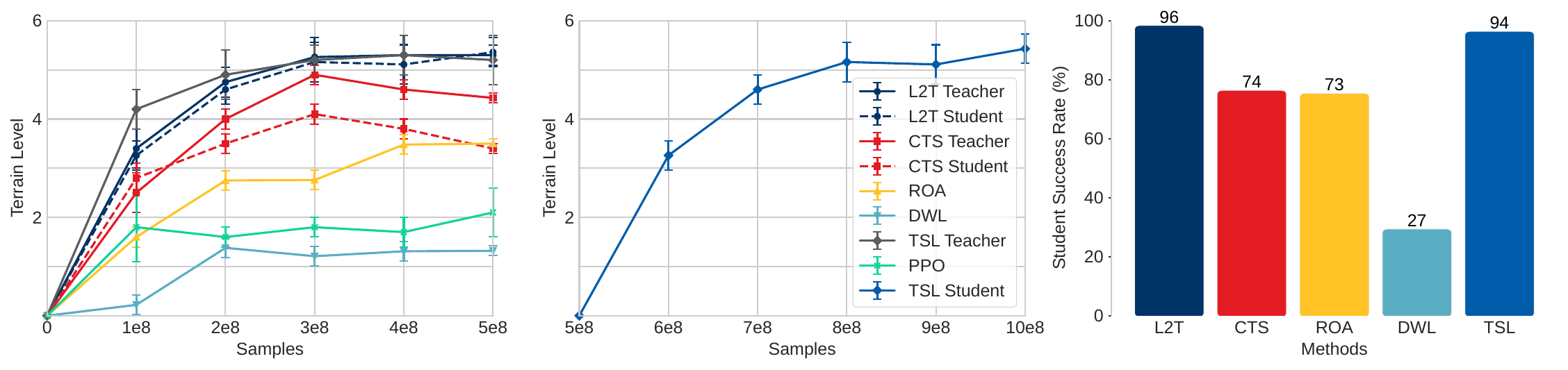}
    \caption{Training curve against baselines. Left: We plot the difficulty of the terrain the agent can successfully traverse during training. Middle: we show the student training curve in TSL after obtaining a teacher. Right: average success rate of the student agents across $4096$ training environments with randomly sampled velocity commands.}
    \label{fig:l2t_terrain_progression}
    \vspace{-0.3cm}
\end{figure*}

\section{Computational Results}
\subsection{Rough terrain difficulty progression}
First, we show the progression of terrain difficulty (curriculum progression) throughout training, representing the overall task completion, i.e., the level of terrain difficulty the agent can walk over with the CoM velocity maintained within a specific range from the velocity command. 
This metric is calculated as the average difficulty level across the $6$ terrain setups mentioned earlier. 
In essence, faster learning corresponds to a steeper curve in terrain difficulty progression. 
Fig.~\ref{fig:l2t_terrain_progression} illustrates the terrain progression during training. 
We represent the teacher policy with a solid blue line and the student policy with a dashed blue line. We compare our approach against three baseline methods based on the asymmetric learning framework: CTS \cite{wang2024cts}, ROA \cite{fu2023deep}, and DWL \cite{gu2024advancing}. In addition, we include comparisons with the conventional teacher-student learning (TSL) paradigm and a vanilla recurrent PPO trained directly in the student’s environment.
Notice that in TSL, the student agent can only be trained after the teacher. The TSL teacher has an MLP-based policy, while the student is LSTM-based, sharing the same network architecture and hyperparameter as in L2T. 
It is noteworthy to mention that we can only obtain a reasonable training result by using data aggregation (DAgger) \cite{ross2011reduction}, i.e., when training the student, periodically use the teacher to predict the action. 

\noindent
\textbf{Discussion:}
First, Fig.~\ref{fig:l2t_terrain_progression} shows that L2T significantly outperforms asymmetric learning baselines, including CTS, ROA, and DWL. We are unable to reproduce competitive performance with DWL due to missing implementation details. However, as DWL follows a similar framework to ROA, learning an encoder to reconstruct privileged states, we believe its potential for improvement over ROA is limited. Since these (single actor) methods lack a separate teacher policy, the student must discover high-reward states with a suboptimal student encoder in order to learn from them, which leads to poor sample efficiency.
Furthermore, we observe a substantial imitation gap between the teacher and student in CTS. This arises because CTS processes teacher and student samples independently, relying solely on a reconstruction loss to align the two encoders. In contrast, L2T mixes samples from both agents, allowing the teacher to observe and evaluate the student’s behavior. From a data-driven viewpoint, CTS distills knowledge using two agents trained on disjoint datasets, while L2T actively injects out-of-distribution student trajectories into the teacher’s training, enabling more effective guidance of the student policy.

Second, compared to the conventional teacher-student learning (TSL) paradigm, we find that L2T enables the student to achieve comparable performance without requiring a separate teacher pretraining stage. That is, the time spent training the teacher in TSL is sufficient for jointly training both teacher and student in L2T, making it a drop-in replacement. While we do not claim that L2T outperforms TSL in final policy quality, we save equivalently $50\%$ training time, or roughly $12$ hours of GPU time, a significant improvement in reducing development cycles. 
Ultimately, both approaches are upper-bounded by the performance of the teacher policy.

\begin{figure}[t]
    \centering
    \subfloat[Imitation gap ablation.\label{fig:imitation_gap}]{
        \includegraphics[width=0.47\linewidth]{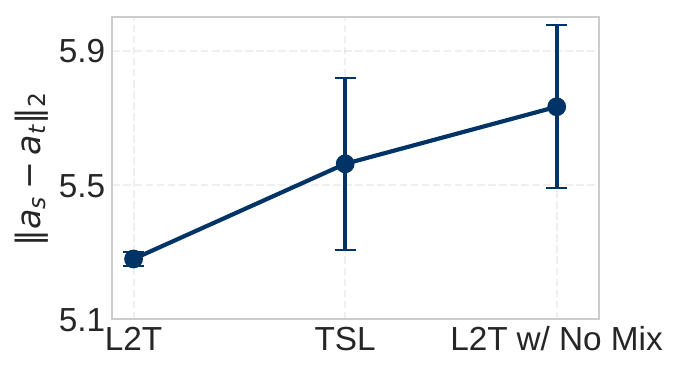}
    }
    \hfill
    \subfloat[Loss function ablation.\label{fig:loss_ablation}]{
        \includegraphics[width=0.47\linewidth]{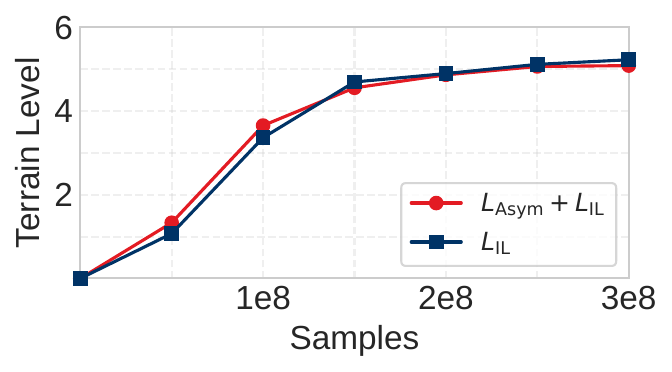}
    }
    \caption{Ablation studies on sample mixing and loss functions.}
    \label{fig:4}
    \vspace{-0.6cm}
\end{figure}
\subsection{Mitigation of imitation gap}
We observe that mixing samples can alleviate the imitation gap, which is caused by the teacher having access to privileged information that is unavailable to the student. 
This privileged information is marginalized during imitation learning for the student agent, resulting in the student agent requiring more exploration and acting more conservatively. 
For example, since the student does not know if it is at the edge of the stairs, they will act less confidently when walking downstairs. 
However, a trained expert teacher agent might act more confidently as it knows the structure of the stairs due to the privileged information of a local depth map.
This is generally true due to the existence of observational noise. 
In the presence of the imitation gap, the teacher agent might generate desirable demonstrations only for the teacher itself, but not necessarily desirable ones for the student agent \cite{weihs2021bridging}. 

In L2T, imitation learning and sample mixing create a bidirectional feedback loop. The teacher policy benefits from student-generated samples, while the student continuously tracks the evolving teacher. This approach resembles techniques that augment training data \cite{shorten2019survey} in supervised learning works such as computer vision, where randomization injects out-of-distribution data into the training set. In the context of our approach, the out-of-distribution data, with respect to the teacher policy, are from the student.
\begin{figure}
    \centering
    \includegraphics[width=0.9\linewidth]{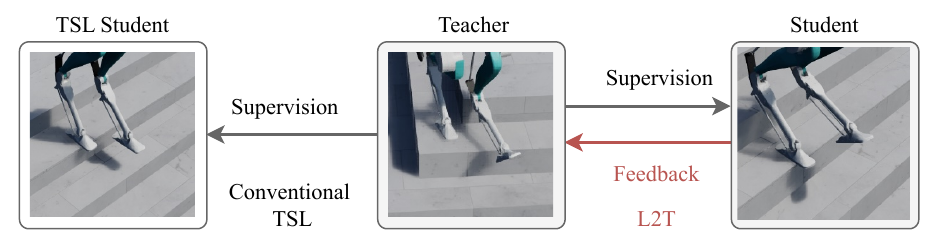}
    \caption{Our method can mitigate the imitation gap, enabling accurately track teacher policy behavior.}
    \label{fig:l2t-divergence}
    \vspace{-0.7cm}
\end{figure}
We show that in Fig.~\ref{fig:l2t-divergence}, the walking gait of the student trained by DAgger is significantly different from the teacher policy, while L2T can faithfully imitate the teacher. Notice the shape of the toe pad from the L2T student, which slightly tilts up from the horizontal plane, accurately mimicking the teacher's toe. In contrast, the DAgger-trained student agent has a flat-ground walking gait, with the toe pad parallel to the horizontal plane. Details of the walking motions can be found in our video.

Additionally, we compute the $L_{\text{IL}}$ between the teacher and student policies under three settings: \textbf{L2T}, \textbf{TSL}, and L2T without sample mixing ($\alpha_{\text{mix}} = 0$) denoted as \textbf{L2T w/ No Mix} as shown in Fig.~\ref{fig:imitation_gap}. This highlights the discrepancy between the two agents across methods and demonstrates that L2T effectively reduces the imitation gap.

\begin{figure}[ht]
  \centering
  \vspace{-0.2cm}
  \includegraphics[width=0.9\linewidth]{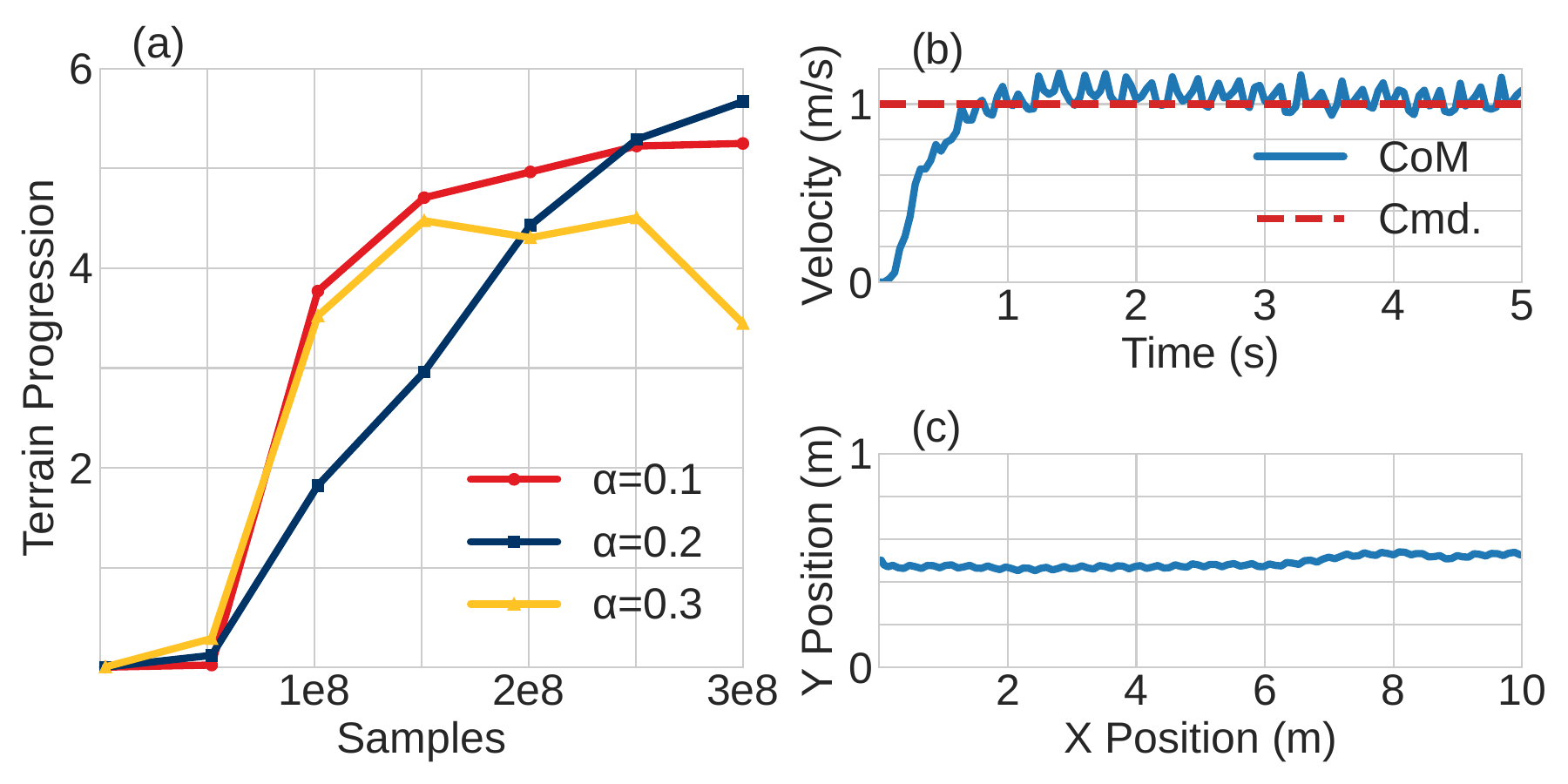}
  \caption{(a) Ablation study on the mixture coefficient. (b) CoM velocity versus command using student policy. (c) $xy$ trajectory with $x$ direction command using student policy.}
  \label{fig:ablation_mc}
  \vspace{-0.5cm}
\end{figure}

\subsection{Ablation study}
We conduct an ablation study on the critical components to further explore key design choices within our algorithm framework. 
Fig.~\ref{fig:ablation_mc}(a) demonstrates the importance of the mixture coefficient $\alpha_{\text{mix}}$ in the training process. 
We observe that while the algorithm becomes unstable with a large $\alpha_{\text{mix}}$, 
which is caused by letting a highly suboptimal student agent inject too many samples,
an appropriate chosen $\alpha_{\text{mix}}$ can benefit the overall training process.
We hypothesize that the discrepancies between the student and teacher promote exploration within the action space, enabling both agents to learn from a broader region around the teacher’s actions. Additionally, we show in Fig.~\ref{fig:loss_ablation} that choosing $L_{\text{IL}}$ slightly outperforms $L_{\text{IL}} + L_{\text{Asym}}$ in our training setup.

\section{Hardware Experiments}
\subsection{Locomotion over real-world uneven terrain}
We report the results of deploying our policy in various outdoor environments, as illustrated in Fig.~\ref{fig:digit_on_terrains} and the supplementary video. These experiments were conducted around a university campus, including 
walkways, wooden bridges, grass hills, beach volleyball courts, and gravel paths.

\begin{figure}[!htpb]
  \vspace{-0.2cm}
  \centering
  \includegraphics[width=0.9\linewidth]{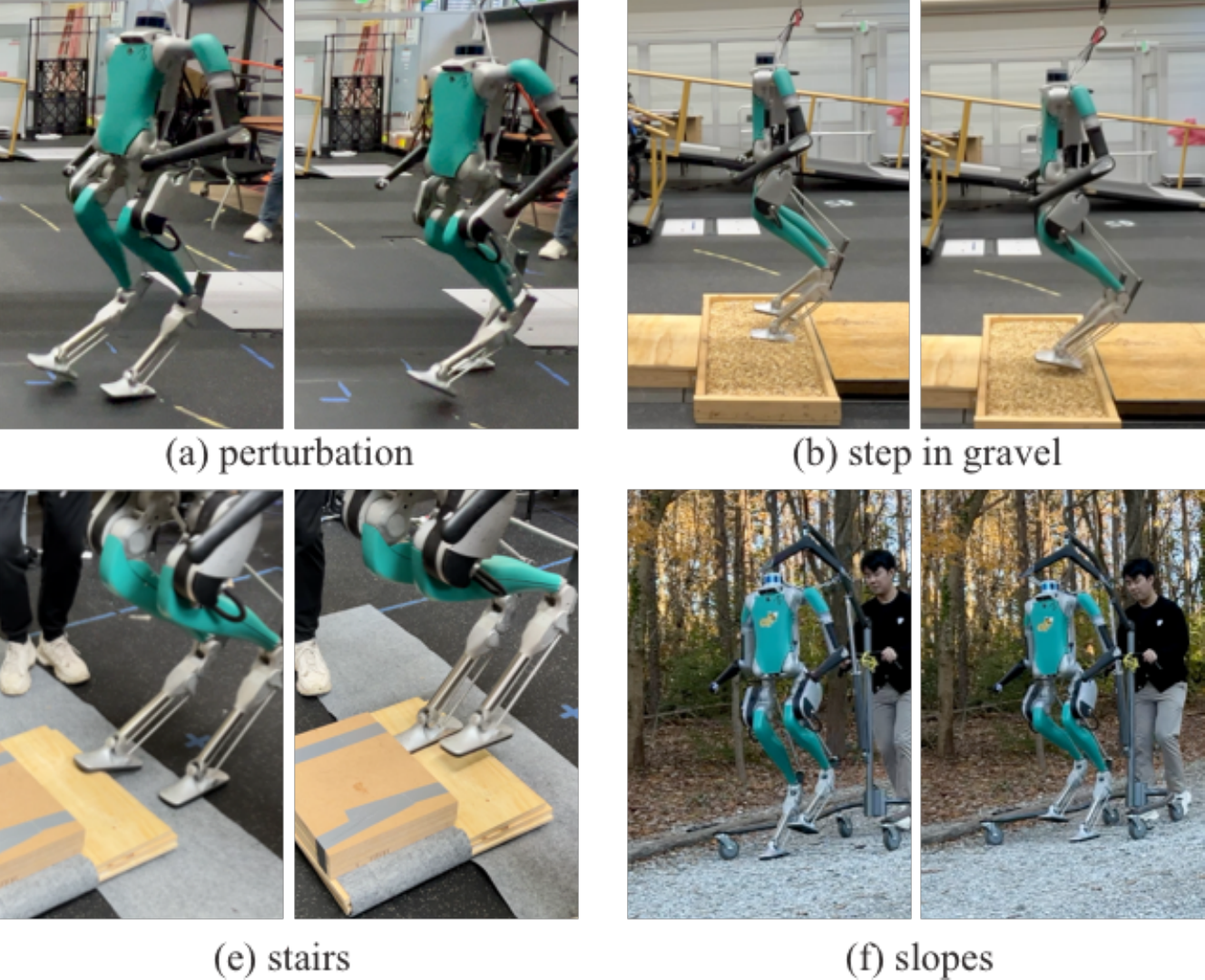}
  \caption{
  (a) we perturb the robot with a harness. (b) we test the step-in-place motion on a crate of gravel. We let the robot walk over  (c) high stairs ($7$cm + $9$cm). (d) hill with gravel.
  }
  \label{fig:perturbation_tests}
  \vspace{-0.5cm}
\end{figure}

Surprisingly, the policy shows generalization ability to scenarios not included in the training. For example, on grass hills, due to rain, the grass and the soil underneath exhibit a certain level of deformation upon impact, which increases the difficulty of state estimation and thus further increases the observation noise. 
Moreover, the policy can walk on the beach volleyball court with sandy terrain, as shown in Fig.~\ref{fig:digit_on_terrains}. 
Despite the robot not being calibrated or trained for such conditions, the policy adapts to the environmental changes without additional training. 
Next, we examine the policy’s performance on terrains with obstacles. The first test involves a crate of gravel shown in Fig.~\ref{fig:perturbation_tests}(b). The robot consistently performs stepping-in-place actions with a stable motion. Even when the robot occasionally strikes the crate's edge, it recovers and resumes normal a stepping gait. 
In the second test, we deploy the robot on a slippery terrain, where we distribute poppy seeds on a whiteboard shown in Fig.~\ref{fig:digit_on_terrains}. Surprisingly, our robot does not exhibit any perceptible shaking motions. In contrast, the company controller provided by Agility Robotics fails to walk over. 

\subsection{Perturbation experiments}
We evaluate the policy’s response to external perturbations during locomotion. Two scenarios are considered: pushing the robot’s center of mass (CoM) from the front, and back using a stick. The policy demonstrates robustness to withstand frontal and rearward pushes while maintaining a stable walking gait.
Additionally, we apply a more substantial perturbation using a harness to pull the robot with an impulsive force (see Fig.~\ref{fig:perturbation_tests}(a)). The robot adapts dynamically, exhibiting agile adjustments to compensate for the pulling.

\section{Conclusion}
We introduced L2T-RL, a novel single-stage learning framework that unifies teacher and student training to address sample inefficiency and enhance real-world performance. Our extensive simulation and hardware experiments demonstrate that L2T-RL achieves robust, and agile locomotion while reducing sample complexity by $50\%$ and thus dramatically saving training time. These results highlight our contributions to redefining teacher-student learning paradigms and paving the way for practical RL-based robotic systems.

\bibliographystyle{IEEEtran}
\bibliography{references}

\end{document}